# Influence of Geometry, Class Imbalance and Alignment on Reconstruction Accuracy – A Micro-CT Phantom-Based Evaluation

Avinash Kumar K M[1], Samarth S. Raut[1*]

[1]Mechanical, Materials and Aerospace Engineering, IIT Dharwad, Dharwad

**Abstract**

The overall accuracy of the three-dimensional (3D) models created from medical scans depends on various factors, including imaging hardware and techniques, segmentation methods, and mesh processing techniques, among others. Although segmentation algorithms exist, the effects of the type of geometry, class imbalance and voxel and point cloud alignment on accuracy remain to be thoroughly explored. This work evaluates the errors across the reconstruction pipeline and explores the use of voxel and surface-based accuracy metrics for different segmentation algorithms and geometry types. Three geometries –a sphere, a facemask, and an abdominal aortic aneurysm (AAA) were printed using the stereolithography technique and scanned using a micro-CT machine. Segmentation was performed using Gaussian Mixture Model (GMM), Otsu thresholding and region-growing (RG) based methods. Segmented and reference models were aligned using the Kabsch-Umeyama (KU) algorithm. They were quantitatively compared to evaluate overlap and classification-based metrics, like Dice and Jaccard scores, precision, among others. Surface meshes, generated using the marching cubes algorithm, were registered with reference meshes using coarse alignment and an Iterative Closest Point (ICP)-based fine alignment process. Surface accuracy metrics like chamfer distance, and average Hausdorff distance were evaluated. The Otsu-based segmentation method was found to be the most suitable method for all the geometries. The AAA model yielded low overlap scores due to its small wall thickness and misalignment. The effect of class imbalance on specificity was observed the most for the AAA model. Surface-based accuracy metrics differed from the voxel-based trends. The RG method performs best for the sphere, while GMM and Otsu perform better for AAA. The facemask surface was most error-prone, possibly due to misalignment during the ICP process. Segmentation accuracy is a cumulative sum of errors across different stages of the reconstruction process. High voxel-based accuracy metrics may be misleading in cases of high class imbalance and sensitivity to alignment. The Jaccard index is found to be more stringent than the Dice score and more suitable for accuracy assessment for thin-walled structures. Voxel and point cloud alignment should be ensured to make any reliable assessment of the reconstruction pipeline.

## 1. Introduction

Computer-aided diagnoses and patient treatments have been seeping into the healthcare infrastructure, augmented by the integration of 3D printing modalities. Medical images have been used for diagnosis since the discovery of X-rays and their successful integration into the clinical diagnostic environment. The accuracy of the final 3D model used for patient-specific treatment depends on every stage of the reconstruction pipeline. The alignment of binary masks generated from original and reconstructed models introduces uncertainty, as the landmark selection process can bias the registration. Due to the inherent bias at every stage, it is not possible to demonstrate an objective reconstruction pipeline that suits all kinds of objects and imaging modalities. Numerous studies, such as those by Kunio [1], [2], Feng et al. [3], Qiang et al. [4], underscore the importance of induction of Computer Aided Diagnoses for improving patient outcomes and strengthening healthcare infrastructure. The use of modelling and simulation tools to assess and frame regulatory practices for the development of medical devices and food substances has received a significant boost [5]. Such computational models have been increasingly used to evaluate the clinical protocols and devices, thereby complementing animal and bench trials [6].

*Corresponding author
Email address: sraut@iitdh.ac.in (Samarth S. Raut)

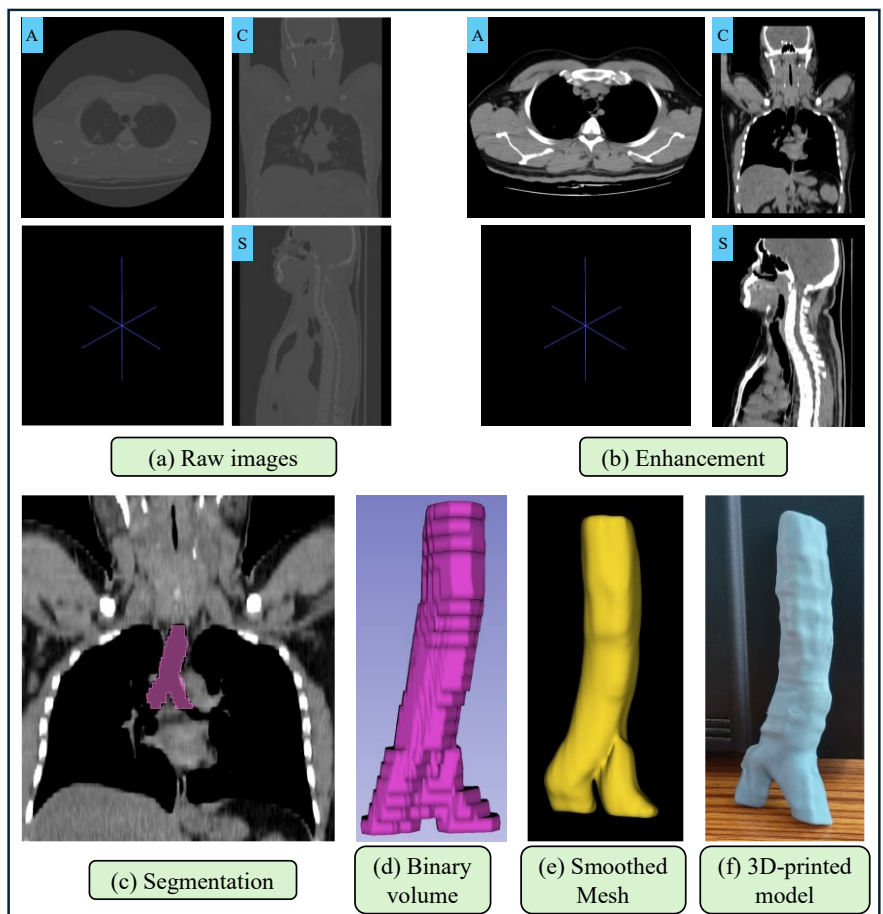

Fig. 1 Illustration of the reconstruction pipeline; A: Axial view, C: Coronal view, S: Sagittal view

Creating 3D physical replicas for medical applications involves image acquisition, enhancement and other image processing techniques, segmentation, mesh generation [7], and 3D printing, each of which introduces its own set of uncertainties, making it challenging to establish a universal workflow relevant to a wide range of geometries. Fig. 1. illustrates an instance of reconstruction of the human trachea leading up to the bifurcation. The original images (a) are contrast-enhanced (b) to enhance delineation between the object and the background. Segmentation creates a 3D model (c), which is then tessellated (d), smoothed using appropriate smoothing algorithms (e) and fabricated (f) using appropriate printing technology.

Several studies have demonstrated the application of the reconstruction pipeline to patient-specific studies and in cases where the ground truth is not available. The imaging modality, imaging parameters, restrictions on the clinically permitted X-ray dosage during imaging, and hardware limitations all affect the quality of scans obtained [8] and introduce errors due to discretization, which manifest as staircase effects [9], as shown in Fig. 2. Material heterogeneity results in partial volume effect [10]. The segmentation technique employed also bases its relevance and efficacy on the imaging technique, heterogeneity in voxel intensity, and local variations in the surface curvature of the object, among other factors. Image processing filters, morphological operations, connected component-based filtering and other operations also influence the segmented models. Table 1 summarizes representative segmentation techniques, their features and typical applications [11], [12].

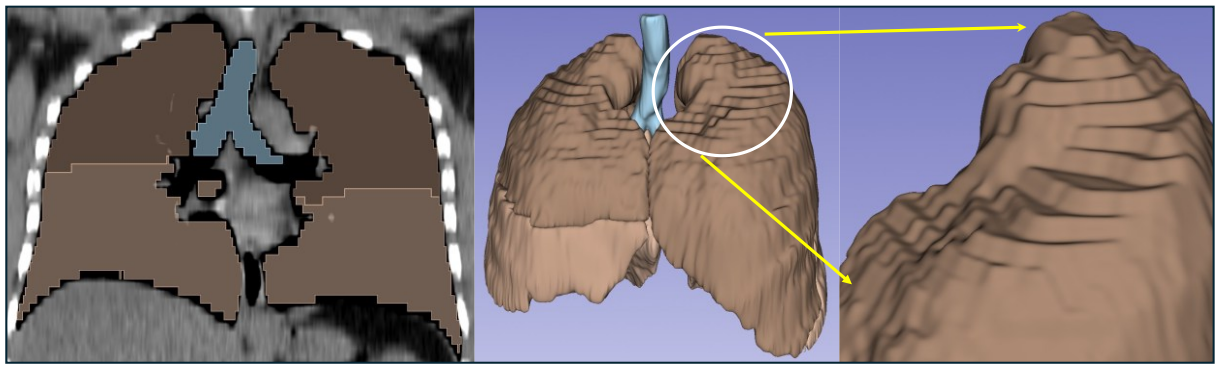

Fig. 2 Staircase effects prominent in one of the segmented lobes of the lungs due to the large slice thickness

The subsequent steps of voxel grid alignment and isosurface extraction are yet another source of error. Surface extraction methods such as marching cubes [13], marching tetrahedra [14], surface nets [15], and dual marching cubes [16] introduce approximation and interpolation errors. The accuracy of tessellated surfaces obtained after surface extraction is also affected by post-processing operations like mesh smoothing. The topology of the object, variations in surface curvature, and the choice of the 3D printing technique may further amplify or mitigate errors.

Table 1 Review of different segmentation techniques

| Technique | Key Features | Advantages | Limitations | Applications |
|---|---|---|---|---|
| Thresholding (Global, Local, Otsu) | Based on grey-level histogram analysis | Simple, fast, easy to implement | Does not consider spatial information, ineffective for intensity inhomogeneities | Bone and tumour CT segmentation |
| Region Growing | Interactive, allows seed points to grow and form regions based on voxel intensity homogeneity | Suitable for connected regions; less noise-sensitive than thresholding | Relies on user input for seed points and may cause over-segmentation | Knee and tissue MRI and CT segmentation |
| k-means | Divides image into k clusters, unsupervised | Simple to implement | Sensitive to outliers and noise | Bone segmentation |

| Edge Detection Based | Boundaries are detected based on discontinuities in pixel values | Accurate boundary localization | Sensitive to noise, used in tandem with region growing to segment | Detection of organ boundaries in CT and MRI images |
| --- | --- | --- | --- | --- |
| Atlas-Based Segmentation | Needs pre-defined anatomical atlases | Accurate and precise for known anatomical models | Ineffective for complex and variable geometries; needs expert knowledge for atlas creation | Brain, liver segmentation |

Many researchers have implemented and assessed the reconstruction pipeline for clinical datasets. Bucking et al. [17] presented a reconstruction pipeline applied to the CT images from the CT MECANIX data set for ribs, liver and lung. The software tools 3D Slicer and Seg3D were used for image segmentation, while the tool MeshMixer was used for mesh processing. Reconstructed models printed were compared with those printed using the Ultimaker 2. The lung exhibited the highest cumulative error of 2.53% while the ribs showed the lowest of 0.78%. Tan et al. [18] developed a soft liver phantom with biliary ducts using 3D printing and moulding methods. The phantom was CT-scanned to get a 512x512x1 voxel grid. The 3D reconstructed models, generated using threshold-based binary segmentation, were compared using the CloudCompare tool. An RMSE of 0.9 +/- 0.2 mm and of 1.7 +/- 0.7 mm for the outer shape of the liver phantom and biliary ducts was reported. Kaschwich et al. [19] reported the semi-automated segmentation and mesh generation using the CTA of six patients. The 3D-printed scanned models reconstructed using Materialise software were compared with the original CT data using centreline analysis, and the mean deviation ranged between -0.73 mm and 0.14 mm. Table 2 highlights the works carried out by various research teams.

While the above studies and those highlighted in Table 2 provide valuable insights, minimal research has been conducted on the impact of different segmentation algorithms on the overall accuracy in the final 3D reconstructed geometry. A systematic investigation addressing the need for voxel and point cloud alignment, as well as the compounded errors resulting from a lack thereof, is not explored. Furthermore, the effectiveness of computer methods for different broad categories of shapes has not been catered to. Addressing this gap, the current study employs micro-CT scanning to investigate a geometric primitive and anatomical models reconstructed using three commonly used segmentation algorithms for three representative geometries - one regular sphere as a reference and two anatomical shapes. GMM, Otsu thresholding and RG based segmentation methods are employed in the current work. Voxel grid alignment is performed using a combination of Principal Component Analysis (PCA)-based landmark identification and the Kabsch Umeyama algorithm, while the point cloud surfaces are aligned using the ICP algorithm. Voxel and surface-based accuracy metrics are evaluated, highlighting their suitability for different variations in geometry and limitations arising from class imbalance and alignment sensitivity.

While this work does not compute the errors arising at every stage of the reconstruction process, it explores the cumulative errors at different stages and underscores the importance of alignment, which may be helpful to other researchers and clinicians looking for a reliable workflow for 3D reconstruction from medical images.

Table 2 Literature review

| Research team | Model/s | Original DICOMs and/or scanning parameters | Segmentation/Reconstruction parameters | 3D printers/ printing parameters | Results/Remarks |
|---|---|---|---|---|---|
| Barker et al. [27] | Dry bone skull and geometric phantom | GE9800 CT scanner, 120 kV, 70 mA, 0.976 x 0.976 x 1.5 mm voxel | Manual thresholding and marching cubes | SLA, with printing time of 20 hours | Mean difference of +0.47 for phantom, absolute maximum, minimum and mean values of +4.62, +0.1, +0.85 mm for skull |
| Pinto et al. [28] | 3 Cadaveric phalanges | 80 kV, 80 mA, 1 mm thick slice, 512 x 512 pixel | Mimics Materialise 13, using thresholding-based segmentation | SLS with in-plane resolution of 0.0875 mm | Local error ranges between -2.7 mm to +3 mm, with errors increasing in regions of high curvature |
| Bilal et al. [24] | 50 mandibular models | Optical white light desktop 3D scanner, 1.31 MP, single-shot accuracy of <= 0.05 mm | 3-matic medical; best fit alignment, global registration | FDM, SLA, SLS, Binder jet | SLS printer showed the highest overall trueness, FDM showed the best precision |
| Dorweiler et al. [23] | 35 aortic and coronary models | 1 mm thick CTA data; 0.26 x 0.26 x 0.4 mm voxel for scanning 3D printed model | Semi-automated thresholding and region-growing methods using Mimics Innovation Suite | FDM and Polyjet printers | 5% increase in wall thickness for all models, mean surface deviation was +0.1 mm for the aortic and +0.18 mm for the coronary models |
| Ravi et al. [26] | 6 anatomical models and 1 geometric primitive | 256x0.625 mm detector row CT, 100-120 kV, 60-100 cc iodine contrast | Manual segmentation | FDM with 0.2 mm layer height | Maximum absolute error of 0.89 mm and maximum relative error of 2.78 % observed across all models |
| Esteve et al. [25] | 3 paediatric liver tumours | 1 mm thick iodinated abdominal CT and T2, DWI and axial post contrast dynamic 3D T1 MRI | Semi-automatic segmentation using IntelliSpace Portal (C) | SLS, FDM | Dimensional error of 3.35, 4.74 and 2.1 mm respectively observed |
| Philip et al. [22] | 11 normal and abdominal AA | 1 mm thick contrast-enhanced CTA | Semi-automated thresholding and region-growing methods using Materialise Mimics | FDM, SLA, SLS and Polyjet printing with layer heights 0.1, 0.05, 0.1, 0.014 mm, respectively | Overall dimensional error of all printing techniques was found to be less than 1 mm |

## 2. Materials and Methods

Three representative models of varying geometric complexities were considered in the current study to assess reconstruction accuracy. Specifically, a sphere, a facemask and an abdominal aortic aneurysm were chosen. The models were fabricated using a Stereolithography SLA 3D printer and imaged using a micro-CT scanner.

### 2.1 Model creation

The CAD model of the sphere designed in SolidWorks® was exported as STEP files. A uniform surface mesh was generated using the open-source meshing tool GMSH [27]. The facemask model was sourced from an open-access repository, while the AAA model was obtained from Raut [28]. Model sizes were selected to fit well within the optimal field of view of the micro-CT scanner. The three models represent an increasing availability of ground truth and a suitable metric for accuracy assessment, and an increasing surface intricacy and local curvature variation, as indicated in Fig. 3.

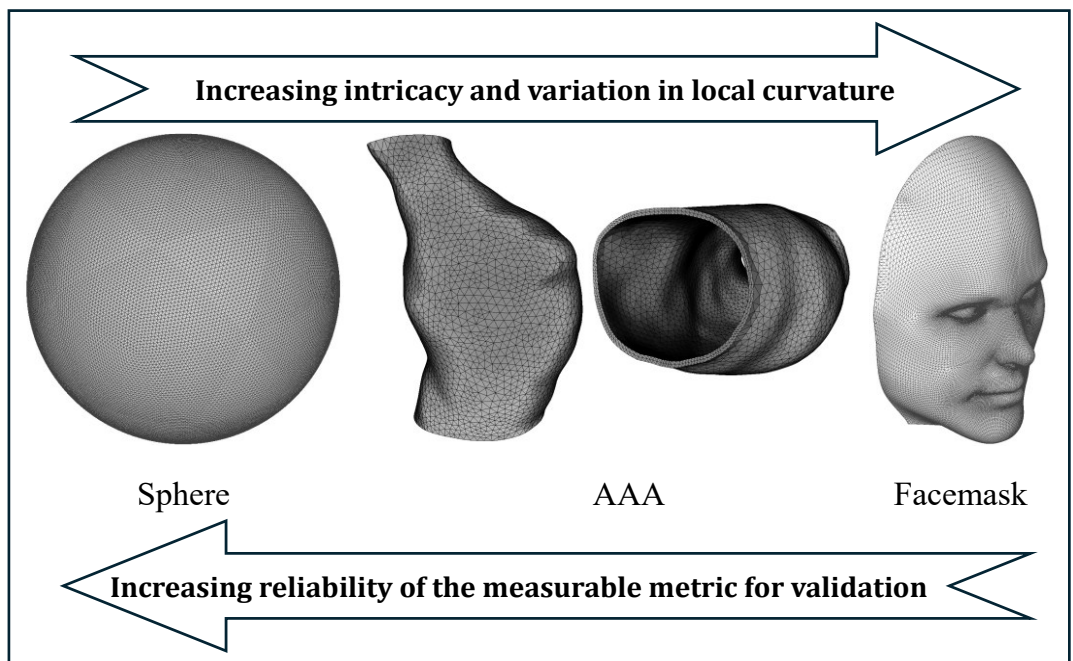

Fig. 3 Geometries used in the current work

### 2.2 3D printing

Models were 3D printed using an SLA technique with a layer height of 25 microns, ensuring the highest available accuracy. Printed models underwent a 10-minute isopropyl alcohol (IPA) bath to remove any vestige of excess resin, followed by UV curing to complete photopolymerization. The printing parameters and post-processing steps are summarized in Fig. 4.

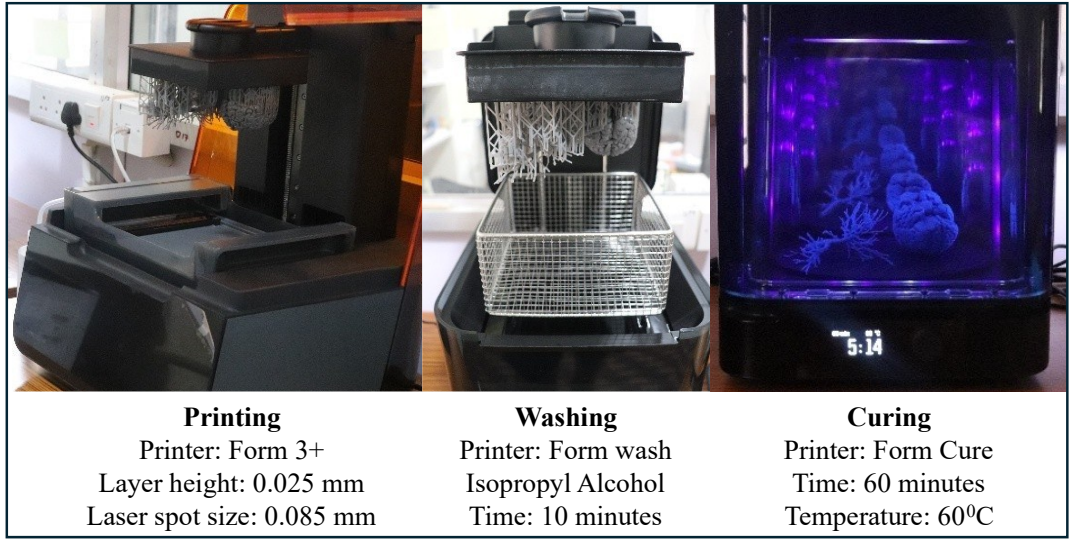

| Printing | Washing | Curing |
|---|---|---|
| Printer: Form 3+ | Printer: Form wash | Printer: Form Cure |
| Layer height: 0.025 mm | Isopropyl Alcohol | Time: 60 minutes |
| Laser spot size: 0.085 mm | Time: 10 minutes | Temperature: 60$^0$C |

Fig. 4 Stages of the model fabrication process using 3D printing

*2.3  Image acquisition*

Micro-CT imaging was performed using Quantum GX2 at the National Facility for Gene Function in Health and Disease (NFGFHD), IISER Pune. Scans were obtained using a 90-kV voltage source with a Field of View (FoV) of 72 mm. The reconstruction FoV was set to 46 mm, resulting in an effective isotropic voxel size of 90 µm in a 512x512x427 voxel grid. Initial scans with a 1.0 mm copper filter lacked adequate sharpness. Hence, a high-resolution scan with a dosage of 1540 mGy/scan was performed without a filter.

*2.4  Reconstruction*

*2.4.1  Segmentation*

A conventional thresholding-based method identifies the region of interest (RoI) by selecting the grey-scale pixel intensity values corresponding to the target material. Manual segmentation is one of the easiest approaches, where the user identifies the values through a trial-and-error approach and/or by using a histogram, which may show a bimodal or multimodal distribution, depending on the number of distinct structures in the scan. Although straightforward, this subjective method may lead to inconsistent results.

To reduce subjectivity, probabilistic clustering models were explored in the current work. A N-component GMM [29] is a weighted sum of densities of Gaussian nature generated using an iterative Expectation Maximization (EM) scheme. GMM generates N sets of means and covariances. The thresholds were computed using these statistical measures.

A representative histogram of pixel intensities of a Digital Imaging and Communications in Medicine (DICOM) slice showing a bimodal Gaussian distribution corresponding to the background (air) and the RoI is shown in Fig. 5. The mean and variance of the Gaussian peak corresponding to the RoI were used to compute the threshold parameters. Another method for selecting values is by Otsu thresholding. It assumes a bimodal distribution of the pixel intensity values and chooses the threshold to minimise the intra-class variance between the two clusters [30].

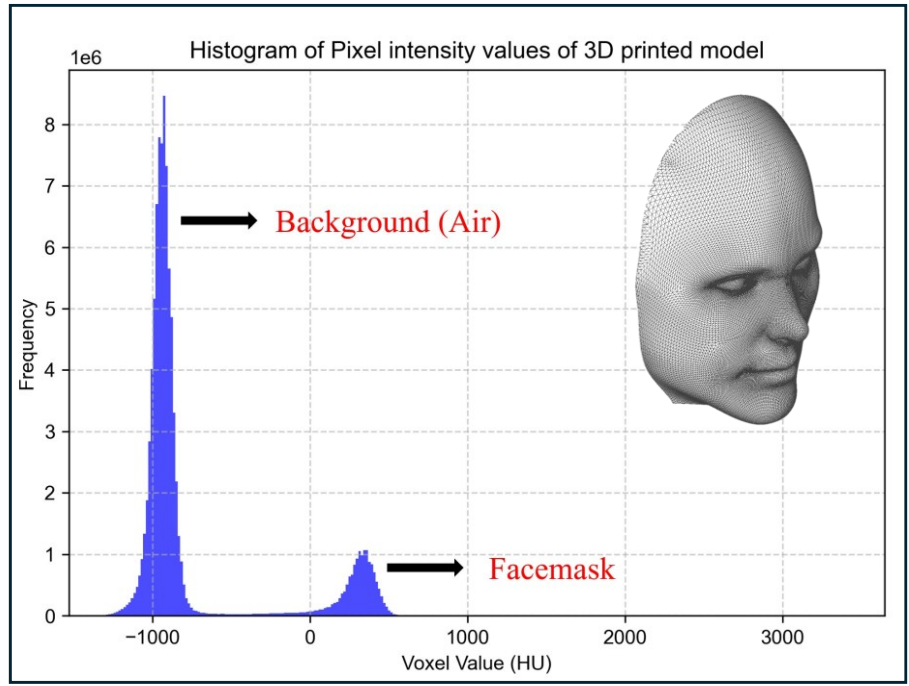

Fig. 5 Bimodal pixel intensity distribution

The RG method is a semi-automatic segmentation technique based on the homogeneity of the pixel intensities. The user-selected seed points initiate region-growing, merging with neighbouring pixels based on similarity until the homogeneity criterion is satisfied to form a distinct region. The accuracy of this semi-automatic method primarily depends on the choice of the seed points and the uniformity of pixel values.

In this study, three types of segmentation procedures were implemented: GMM-based, Otsu thresholding, and RG-based. GMM-based and Otsu-based segmentation methods were implemented in Python using open-source packages like scikit-learn [31] and scikit-image [32], while the region-growing method was performed using the open-source tool ITK-SNAP [33].

*2.4.2    Voxel grid alignment and mesh generation*

The reference models, initially represented as surface tessellations, were converted into voxel grids with spacing corresponding to the slice thickness and pixel spacing of the DICOM data obtained during the image acquisition stage. An initial coarse alignment was performed using a combination of manual translation and rotation to facilitate accurate correspondence between reference and reconstructed voxel grids. This was followed by fine alignment using the KU algorithm [34], which computes the optimal rotation and translation between the pairwise voxels on the reference and reconstructed models. Also called landmark points, these pairwise voxels are typically defined based on salient features of the models. Traditional landmark selection methods, such as those using the corners of bounding boxes, convex hulls, or principal grids, are grid-dependent and sensitive to the initial orientation of the object in voxel space. To overcome this limitation, a robust, grid-independent landmark identification approach based on PCA [35] was implemented.

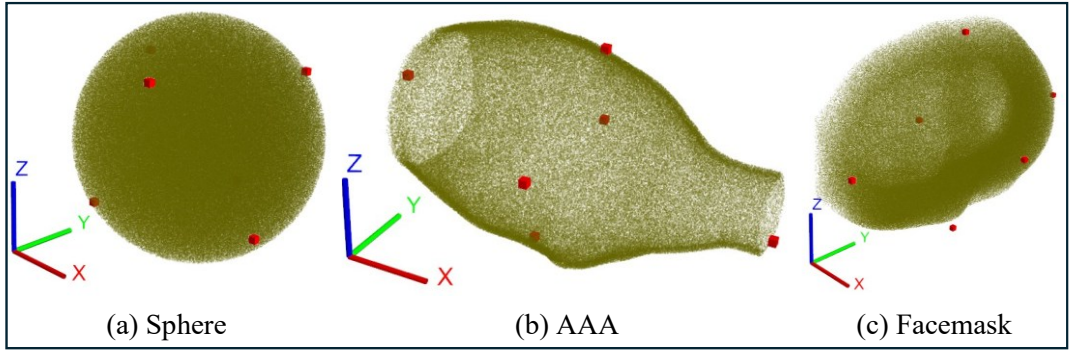

(a) Sphere  (b) AAA  (c) Facemask

Fig. 6 Landmarks (red points) identified by a grid-independent PCA-based method for three different geometries, namely Sphere, AAA and facemask

In the PCA-based method, the centroids of the two point-sets were aligned to obtain the translation vector. The covariance matrix of the centred data was decomposed to obtain the principal directions. Projecting the voxels along with these directions gave the extremal directions and the landmarks corresponding to the principal directions, as shown in Fig. 6.

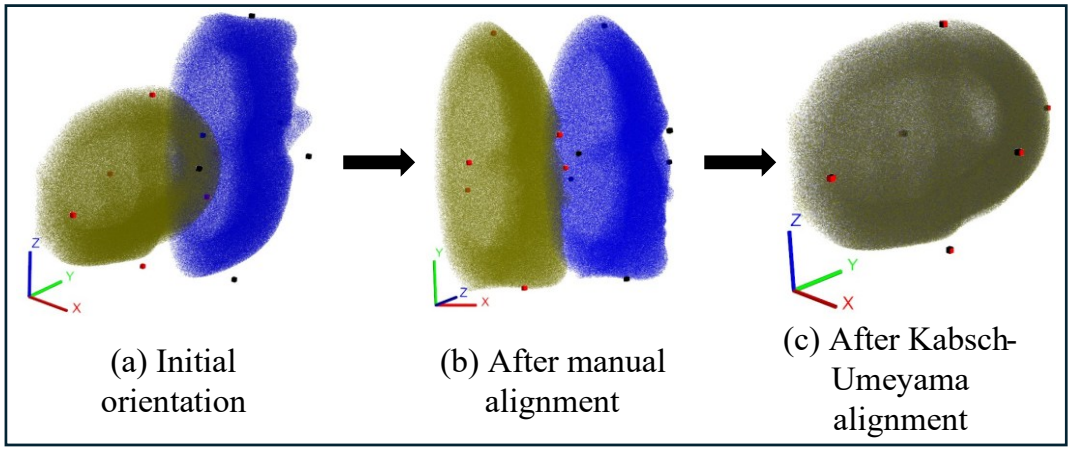

(a) Initial orientation  (b) After manual alignment  (c) After Kabsch-Umeyama alignment

Fig. 7 Process of alignment of voxel grids using the Kabsch-Umeyama algorithm for the facemask

The implementation of the Kabsch-Umeyama algorithm is shown in Fig. 7. The manually aligned reconstructed binary voxel grid provided a good initialization point for the KU algorithm. The rotation and translation matrices computed using the KU were used to align the centroids, followed by the rotation of the landmarks and the full voxel grids, respectively, ensuring accurate alignment of the reconstructed voxel grid.

Following the alignment, isosurface extraction was performed to generate surface meshes. The marching cubes algorithm was used in the current study.

### 2.4.3 Point cloud alignment

The reference and reconstructed surface meshes were registered using a two-stage pipeline using the Open3D library [36]. The surface normals of the corresponding two point clouds were downsampled to a sufficient degree, reducing computational complexity while maintaining geometric fidelity. Fast Point Feature Histograms (FPFH) [37] were computed for each downsampled point cloud to perform feature-based matching. Global registration by coarse alignment was performed by Random Sampling Consensus (RANSAC)-based feature matching. This was followed by ICP [38] point-to-plane registration. The resulting transformation computed after convergence was applied to the reconstructed model to achieve a robust and accurate alignment. This alignment protocol was followed for all the reconstructed models generated using different segmentation algorithms.

### 2.5 Accuracy metrics

Binary segmentations in the current work are assessed by computing classification and overlap based metrics such as sensitivity, specificity, precision, Dice Index (DI) and Jaccard Index (JI). Proper voxel grid alignments were ensured as a crucial prerequisite for meaningful interpretations. Volume similarity, which is independent of the alignment process also provides a measure the correctness of segmentation. The local surface deviations are assessed using the metrics such as Chamfer Distance (CD), Average Hausdorff Distance (AHD) and Root Mean Square Error (RMSE). More information about the accuracy metrics is found in the Supplementary section.

### 2.6 Post-processing operations

Mesh smoothing involves repositioning the vertices of a mesh according to a specific rule to minimise the undulations and jaggedness introduced by discretisation and interpolation errors during image acquisition and surface extraction. Laplacian smoothing is one of the simplest and easiest-to-implement smoothing algorithms. It traverses all the vertices and moves them to new locations driven by their respective neighbours, and a scaling factor regulates the degree of smoothness achieved by each iteration.

The overall reconstruction and comparison workflow – from model creation to accuracy assessment is summarized in Fig. 8.

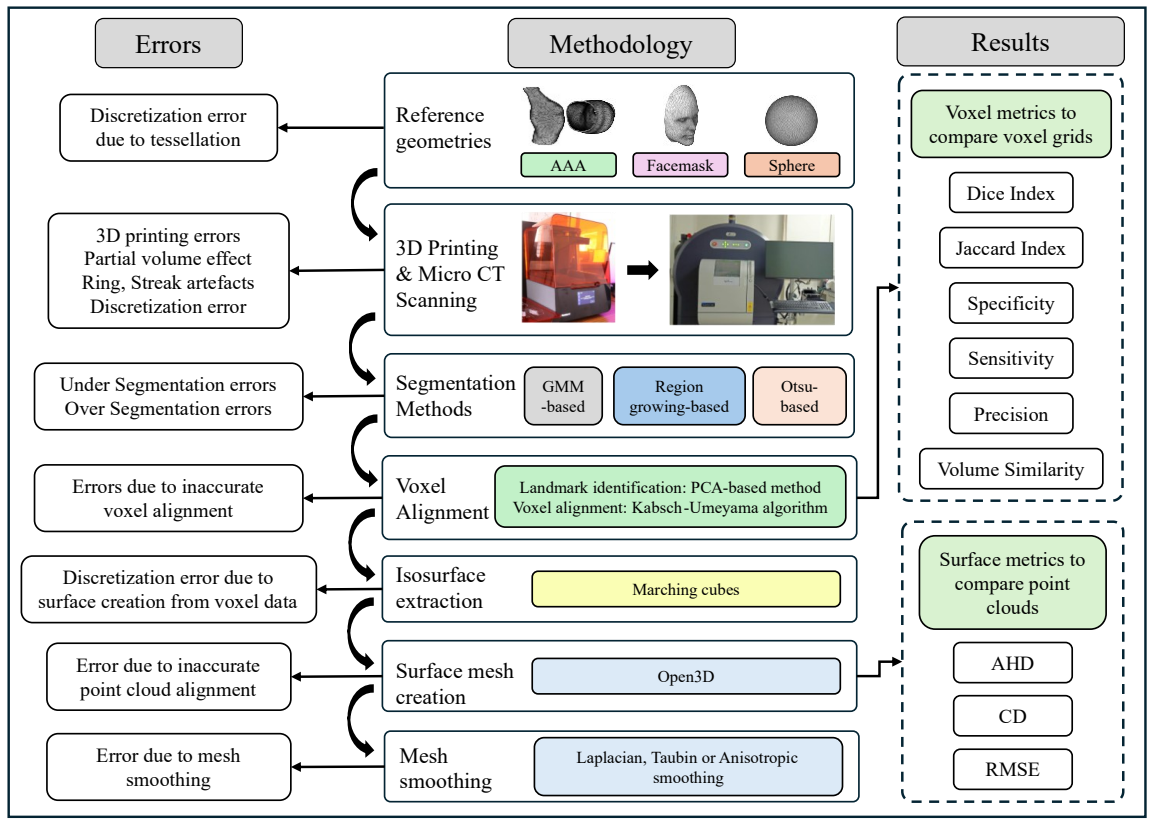

Fig. 8 Methodology employed in the current study

# 3. Results

## 3.1 Foreground voxel distribution

The percentage of voxels representing the given binarized voxelised reference geometry in a standard 512x512x427 voxel grid is shown in Fig. 9. 10.021% and 10.492% of the total voxels are occupied by the sphere and facemask geometries, respectively, while the AAA model is represented by only 0.983% of the total voxels. This evident difference indicates a strong class imbalance for the AAA model, and that the classification-based metrics may behave differently for different geometries.

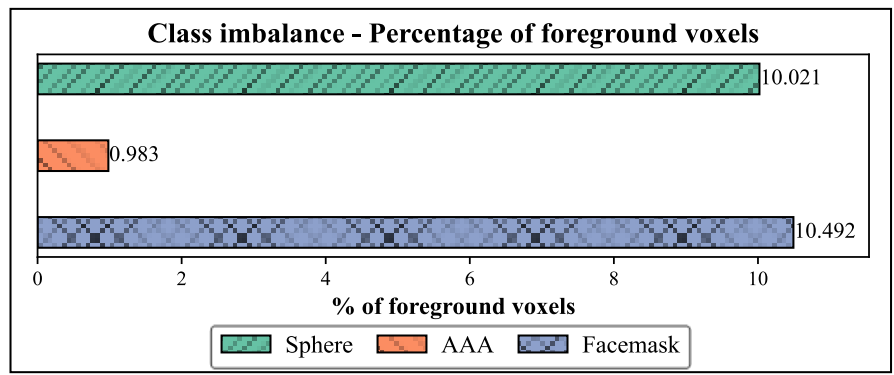

Fig. 9 Percentage of foreground voxels occupied by the reference geometries in a standard (512x512x427) voxel grid

## 3.2 Overlap-based voxel accuracy

The DI and JI scores computed for each geometry and segmentation method are shown in Fig. 10.

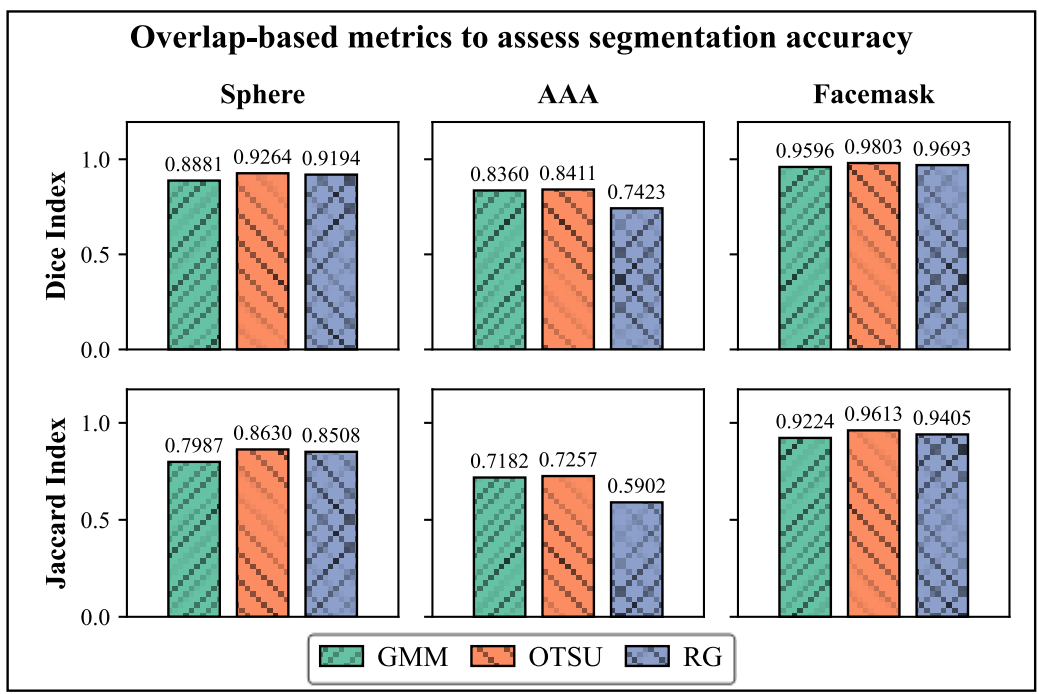

Fig. 10 Overlap-based metrics comparing the accuracy of segmented binary voxel grid output of three segmentation techniques explored for three types of geometries

The Otsu method exhibits the highest DI and JI scores consistently for all geometries, with DI-JI scores of 0.9803-0.9693 for facemask, 0.92614-0.8630 for sphere and 0.8411-0.7257 for AAA models. However, the GMM method resulted in lowest DI-JI values for sphere (0.8881-0.7987) and facemask (0.9596-0.9224). The AAA model was found to be most sensitive to errors due to the thin-walled nature of the geometry and misalignment, irrespective of the segmentation method used. The RG method yielded the least DI-JI values of 0.7423-0.5902 for the AAA model.

*3.3   Volume similarity*

Volume similarity, in contrast to the overlap indices, is independent of the alignment of the voxel grids and is purely a result of the segmentation process. The Otsu method creates models that exhibit the highest similarity to the respective voxel grids for all segmentation methods, as shown in Fig. 11. The facemask model shows the highest similarity (0.993), followed by the AAA model (0.992) and sphere (0.936). The GMM method performs worst for the facemask and sphere models (0.963, 0.890), while the RG method yields the highest dissimilarity for the AAA model.

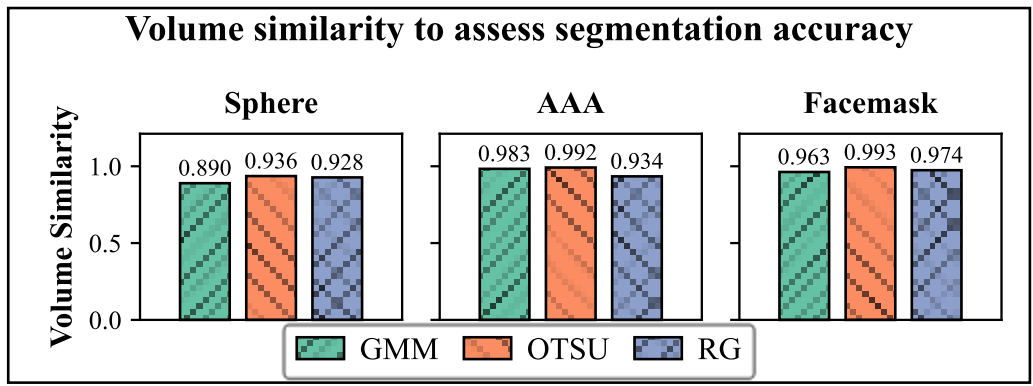

Fig. 11 Volume similarity between reference and segmented binary voxel grids

*3.4   Classification-based accuracy metrics*

The classification-based metrics computed for each geometry and segmentation method are presented in Fig. 12. The Otsu method resulted in the highest sensitivity values for all geometries, with a maximum for the facemask (0.9737), followed by the sphere (0.8709) and AAA model (0.8480). The GMM method performed poorly for the sphere (0.7999), while the RG method yielded the lowest sensitivity value for AAA (0.7945). The sphere and facemask models resulted in high precision values for all the segmentation methods. The AAA model showed the least precision for all segmentation methods, with the RG method performing the worst (0.6965) and the GMM method being relatively best (0.8502). All the segmentation methods yielded voxel grids with very high specificity, exceeding 0.996 across all geometries. This is attributed to the dominance of background voxels, as evidenced by the percentage of foreground voxels computed in section 3.1. These trends show the influence of class imbalance and voxel alignment on specificity and precision.

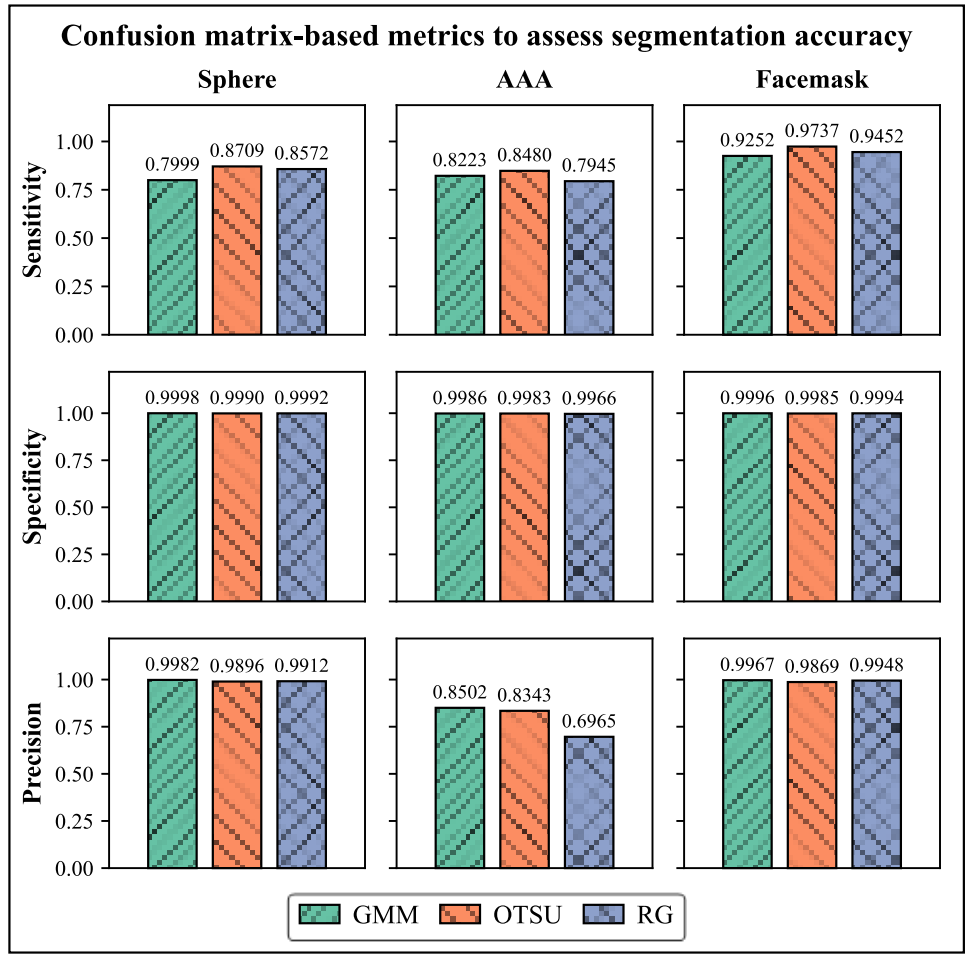

Fig. 12 Confusion matrix (Classification)-based metrics comparing the accuracy of segmented binary voxel grid output of the segmentation techniques explored for the three geometries

### 3.5 Surface-based accuracy metrics

The assessment of surface accuracy using CD, AHD and RMSE is illustrated in Fig. 13. The RG method performs best for the sphere model, with the least chamfer distance (0.0075 mm), followed by the Otsu and GMM methods (0.0077 mm and 0.0099 mm, respectively). The surfaces produced by the AAA method are accurate to a similar degree, with chamfer distances of 0.0177 mm for the GMM and Otsu methods and 0.0124 mm for the RG method. On the other hand, the tessellation of the facemask produced by the GMM method is the least accurate (0.0665 mm), followed by those produced by the RG and Otsu methods (0.014 mm and 0.0097 mm, respectively). A similar pattern is observed for the AHD, with RG performing the best for the sphere (0.0063 mm), followed by the Otsu and GMM methods (0.0064 mm and 0.0093 mm, respectively). The AAA surfaces produced by all three methods have similar accuracy (0.011 mm, 0.011 mm, and 0.0017 mm for the GMM, Otsu, and RG methods, respectively). The facemask surface with the least AHD is produced by the Otsu method (0.0085 mm), followed by the RG and GMM methods (0.0121 mm and 0.0455 mm, respectively). The RMSE of surface deviations of the facemask model produced by the GMM method are highest (0.0853 mm), followed by the RG and Otsu methods (0.0141 mm and 0.0096 mm, respectively). The spherical surfaces produced by the Otsu and RG methods have an RMSE value of around 0.007 mm, while the GMM method performs the worst (0.01 mm).

AAA surfaces produced by GMM and Otsu have an RMSE value of 0.0116 mm, while the RG method produces the least accurate surface (0.0123 mm).

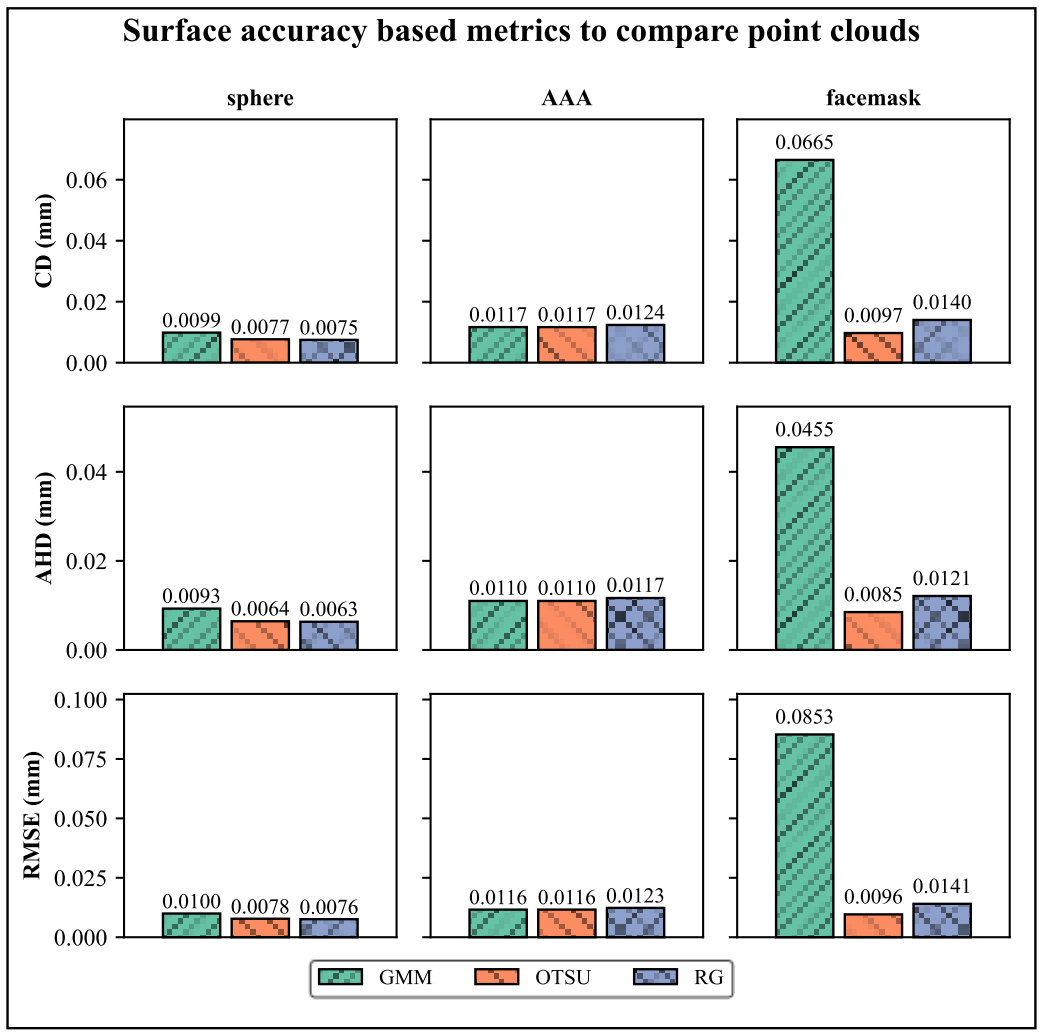

Fig. 13 Surface error metrics comparing the accuracy of surface tessellations

## 4. Discussions

The anatomical models, facemask, and AAA, along with a standard geometry sphere, were considered in the current study. Patient-specific clinical CT or MRI data are inherently heterogeneous and complex, making them challenging to segment with high accuracy and fidelity. However, to minimise additional uncertainty and case-to-case operator-specific subjective variations, our study focuses on the operational effectiveness of algorithmic components, using deliberate and consistent material, a 3D printer, and a micro-CT scanner for all three representative cases.

### 4.1 Influence of geometry and class imbalance

The significant disparity between the percentage of foreground voxels across the three geometries strongly influenced the classification-based accuracy metrics. The AAA model, which occupied less than 1% of the total voxels, showed high specificity but low values of precision and sensitivity, especially for the RG method. This behaviour is expected in the instances where the background voxels dominate, and even a small error in segmentation has minimal effect on specificity but a significant impact on sensitivity and precision. The sphere and facemask, being relatively bulky volumes with at least 10% of total voxels representing them, show stable precision and sensitivity. It is worth noting that the segmentation accuracy metrics should be interpreted by considering the presence or absence of class imbalance, particularly in cases involving thin-walled anatomical geometries. This inference corroborates well with the findings by Mueller et al. [39].

*4.2 Sensitivity of voxel accuracy metrics to alignment*

The AAA model exhibited a strong correlation between voxel-grid alignment and overlap-based metrics. Even though the Otsu method resulted in very high volume similarity, the DI and JI values left to be desired, especially for AAA model. The bulkier models – sphere and facemask – were less sensitive to the error caused by misalignment. Because the AAA model has a very small wall thickness (~0.5 mm), a small misalignment during the KU algorithm can cause several voxel rows and columns to shift, reducing the overlap. The effect of these inaccuracies is more pronounced on JI compared to DI. For AAA, when segmented using Otsu, the Dice score decreases from 0.8411 to 0.7257, representing a decrease of 0.1154. In contrast, the Jaccard index decreases from 0.7423 to 0.5902, showing a decrease of 0.1521. The Jaccard index was found to be more punishing than the Dice index. This is in line with the findings reported by Taha et al. [40]. Thus, to accurately assess the quality of segmentation and the effect of alignment, JI provides a more stringent measure of accuracy.

*4.3 Geometry-specific behaviour of surface metrics*

The tessellations resulting from the marching cubes algorithm are compared with reference meshes using a combination of RANSAC-based feature mapping and ICP-based point-to-plane mapping. Due to its constant curvature and minimal surface variations resulting from over- and under-segmentation, the sphere surface exhibited the highest surface accuracy, as indicated by very low values of CD, AHD, and RMSE of surface deviations.

The lower values of DI, JI and precision for AAA, which were due to small wall thickness and slight increase in surface variations, were mitigated by the point cloud alignment procedure employed in the current study. However, the facemask surface with the highest local surface variations suffered the most from misalignment. The flat surface of the distal end of the facemask, coupled with local surface variations on the proximal side, corresponding to anatomically rich regions (such as the nose, lips, and eyes), posed a challenging problem for the coarse and fine ICP alignment procedure. This led to a slight increase in the values of CD, AHD, and RMSE of 0.065 mm, 0.0455 mm, and 0.0853 mm, respectively, especially for the facemask surface, resulting from the GMM segmentation method.

*4.4 Metric selection in medical image segmentation*

The results expound the importance of selecting the accuracy metrics based on the geometry being considered. For thin-walled geometries and sparsely represented structures, surface-based metrics and JI provide more reliable evaluations, rather than using DI alone. For the geometries with a balanced voxel distribution, the voxel and surface accuracy metrics are in agreement. However, the surface misalignment due to the local curvature-driven changes for the facemask surface demonstrates that one should not rely solely on a single metric for accuracy assessment.

*4.5  Error propagation in the reconstruction pipeline*

The reconstruction pipeline is a process that contains multiple and interdependent steps – segmentation, voxel-grid alignment, surface extraction, point cloud alignment and 3D printing. Each stage produces errors that percolate into subsequent stages, which may amplify or mitigate the overall errors. Similar observations are reported by George et al. [41].

The overlap and classification-based accuracy metrics do not provide a measure of just the segmentation accuracy, but rather a cumulative sum of errors accumulated across all the stages preceding it. Similar remarks can be made for the point cloud accuracy assessment metrics as well. It is challenging to measure error at each stage without introducing an error in the measurement process itself. The present study did not isolate the error contribution of each stage but highlights the cumulative nature of the reconstruction error and the need for robust, geometry-aware alignment procedures.

Overall, this work presents a working reconstruction pipeline and guidelines for selecting segmentation metrics based on the type of geometry and class imbalance. It also highlights the importance of alignment in assessing 3D reconstructions. Future work will include geometries spanning a wide range with larger variations in surface curvature, as well as implementing machine learning-based segmentation and accuracy assessment. Controlled perturbations to quantify the sensitivity of the alignment process and their impact on assessment can be explored.

**Credit authorship contribution statement**

Avinash Kumar K M: Coding, Writing – review & editing, Writing – original draft, visualization, methodology, conceptualization, formal analysis

Samarth S. Raut: Writing – conceptualization, review & editing, data curation, formal analysis

**Declaration of competing interest**

The authors declare that they have no known competing financial interests or personal relationships that could have appeared to influence the work reported in this paper.

**Ethics statement**

This study did not involve any experiments on human participants or animals. No ethical approval or informed consent was required.

**Funding**

The work was supported by Govt. of India SERB Start-Up Research Grant SRG/2020/002513 and BITS BioCyTiH Foundation grant.

**Glossary**

AAA: Abdominal Aortic Aneurysm

GMM: Gaussian Mixture Model

RG: Region Growing

KU: Kabsch-Umeyama

ICP: Iterative Closest Point

RMSE: Root Mean Square Error

PCA: Principal Component Analysis

SLA: Stereolithography

SLS: Selective Laser Sintering

FDM: Fused Deposition Method

IPA: Isopropyl Alcohol

FoV: Field of View

DICOM: Digital Imaging and Communications in Medicine

FPFH: Fast Point Feature Histograms

RANSAC: Random Sampling Consensus

DI: Dice Index

JI: Jaccard Index

CD: Chamfer Distance

AHD: Average Hausdorff Distance

## Supplementary material

**Accuracy metrics**

Binary segmentation in the current work can be assessed by performing classification-based performance analyses, which generate the four usual suspects: True Positives (TP), True Negatives (TN), False Positives (FP), and False Negatives (FN). Using these four metrics, Sensitivity, Specificity and Precision are derived.

$$\text{Sensitivity (Recall)} = \frac{TP}{\text{Actually Positive}} = \frac{TP}{TP + FN} = \text{True Positive Rate}$$

$$\text{Specificity} = \frac{TN}{\text{Actually Negative}} = \frac{TN}{TN + FP} = \text{True Negative Rate}$$

$$\text{Precision} = \frac{TP}{TP + FP} = \text{Positive Predictive value}$$

Volume similarity is another metric that is independent of orientation and compares the volume of the two voxel grids.

$$\text{Volume similarity} = 1 - \frac{|FN - FP|}{2 * TP + FP + FN}$$

Suppose $V_{ref}$ and $V_{recon}$ are the voxel grids corresponding to the reference (original) model and the reconstructed model.

a. Dice index (DI): It measures the overlap between the two voxel grids and is dependent on the final alignment of the $V_{ref}$ and $V_{recon}$.

$$\text{DICE} = \frac{2 * |V_{ref} \cap V_{recon}|}{|V_{ref}| + |V_{recon}|}$$

It ranges from 0 (no overlap) to 1 (perfect overlap)

b. Jaccard index (JI): Also known as Intersection over Union (IOU), this is a stringent measure of overlap between the two voxel grids.

$$\text{Jaccard} = \frac{|V_{ref} \cap V_{recon}|}{|V_{ref} \cup V_{recon}|}$$

The Jaccard index is mathematically equivalent to the F1 score, which is used in assessing the classification accuracy of machine learning models.

Local surface deviation and distance-based metrics were computed to quantify the accuracy of each reconstructed mesh compared to the corresponding reference model. The following geometric accuracy metrics were evaluated:

i. Root Mean Square Error (RMSE): RMSE is a global deviation metric that quantifies the average magnitude of the surface deviations between corresponding points on the original and reconstructed mesh.

$$\text{RMSE} = \sqrt{\frac{1}{N}\sum_{i=1}^{N} d_i^2}$$

Where,

$d_i^2$ is the Euclidean distance between the $i^{th}$ corresponding point between the two meshes, and $N$ is the minimum of the total number of points in the two point clouds.

In the instances where one-to-one mapping between the two meshes was not possible, an octree-based nearest neighbour implementation for closest point mapping was followed, where for each point on the reconstructed mesh ($p_i$), the closest point on the reference mesh ($q_i$) is identified, and the Euclidean distance ($d_i = \|p_i - q_i\|$) between such pairs is considered. The point clouds were spatially aligned, as explained in Section 2.4.3.

i. Average Hausdorff distance (AHD): Hausdorff distance represents the maximum surface deviation between the segmented volume and ground truth. This metric is sensitive to outliers present in either of the point clouds and is not recommended. Average Hausdorff Distance, which is the HD averaged over all points, is utilized.

$$AHD\ (A, B) = avg\left(\frac{1}{A}\sum_{a \in A} \min_{b \in B} d(a, b) + \frac{1}{B}\sum_{b \in B} \min_{a \in A} d(a, b)\right)$$

ii. Chamfer distance: It is the average of the sum of Euclidean distances of point sets A to B and vice versa.

$$CD\ (A, B) = \frac{1}{|A|}\sum_{a \in A} \min_{b \in B} \|a - b\|_2^2 + \frac{1}{|B|}\sum_{b \in B} \min_{a \in A} \|b - a\|_2^2$$

Where,

$A$ and $B$ are the two point sets, $a$ and $b$ are the points on the respective point sets.